\documentclass{bmvc2k}

\usepackage{microtype}
\usepackage{graphicx}
\usepackage{subfigure}
\usepackage{booktabs}
\usepackage{multirow}
\usepackage{multicol}
\usepackage{amsmath}
\usepackage{amssymb}
\usepackage{mathtools}
\usepackage{amsthm}
\usepackage{tabularx}
\usepackage{wrapfig,lipsum,booktabs}
\usepackage{enumitem}
\usepackage[symbol]{footmisc}
\title{Class-Continuous Conditional Generative Neural Radiance Field}

\addauthor{Jiwook Kim}{tom919@cau.ac.kr}{1}
\addauthor{Minhyeok Lee*}{mlee@cau.ac.kr}{1}

\addinstitution{
 School of Electrical \& Electronics Engineering\\
 Chung-Ang University\\
 Seoul 06974, Republic of Korea
}

\runninghead{Jiwook Kim and Minhyeok Lee}{$\textit{C}^{3}$G-NeRF}

\def\etal{\emph{et al}\bmvaOneDot}

\begin{document}

\maketitle

\begin{abstract} 
The focus of 3D-aware image synthesis lies in preserving spatial consistency while generating high-resolution images with fine details. Recently, Neural Radiance Field (NeRF) has emerged as a powerful method for synthesizing novel views with low computational cost and exceptional performance. Although existing generative NeRF approaches have achieved significant results, they are unable to handle conditional and continuous feature manipulation during the generation process. In this work, we present a novel model, called Class-Continuous Conditional Generative NeRF ($\textnormal{C}^{3}$G-NeRF), which synthesizes conditionally manipulated photorealistic 3D-consistent images by projecting conditional features onto the generator and discriminator. We evaluate the proposed $\textnormal{C}^{3}$G-NeRF on three image datasets: AFHQ, CelebA, and Cars. Our model demonstrates robust 3D-consistency, fine details, ability of 360$^{\circ}$ generation, and smooth interpolation in conditional feature manipulation. For example, $\textnormal{C}^{3}$G-NeRF achieves a Fréchet Inception Distance (FID) of 7.64 in 3D-aware face image synthesis with a $\textnormal{128}^{2}$ resolution. Furthermore, we provide FIDs and for generated 3D-aware images of each class within the datasets, showcasing the ability of $\textnormal{C}^{3}$G-NeRF to synthesize class-conditional images.
\end{abstract}

\section{Introduction}
\label{introduction}
There have been many approaches \cite{VAE, pixel_rnn} for synthesizing novel images. Generative Adversarial Network (GAN) \cite{gan} has shown outstanding results in generating images by learning the distributions of datasets, resulting in the synthesis of photo-realistic instances. Furthermore, many studies have improved the ability to generate high-resolution images \cite{chen2016infogan, karras2017progressive, karras2020analyzing, choi2018stargan}. Notwithstanding the advances made by existing studies, GANs still have limitations when synthesizing multiple views of a single object due to most data collections being based on two-dimensional information, which can lead to instability in 3D-consistency.

In order to address this problem, 3D-based GANs \cite{wu2016learning, nguyen2019hologan, nguyen2020blockgan} have been studied, which make use of volume rendering methods. However, those methods require high computational power and memory in order to train effectively. In recent years, Mildenhall \etal{} \cite{mildenhall2021nerf} proposed the Neural Radiance Field (NeRF) as an alternative to conventional voxel-based volume rendering methods \cite{kajiya1984ray, drebin1988volume, henzler2019escaping}. NeRF greatly reduces the complexity of computations and memories required compared to other approaches. Accordingly, NeRF has been extended for use in 3D-based GAN studies \cite{schwarz2020graf, niemeyer2021giraffe, chan2021pi, deng2022gram} with excellent results and low complexities.

\begin{wrapfigure}{r}{0.5\textwidth} 
\begin{center}
\centerline{\includegraphics[width=0.5\columnwidth]{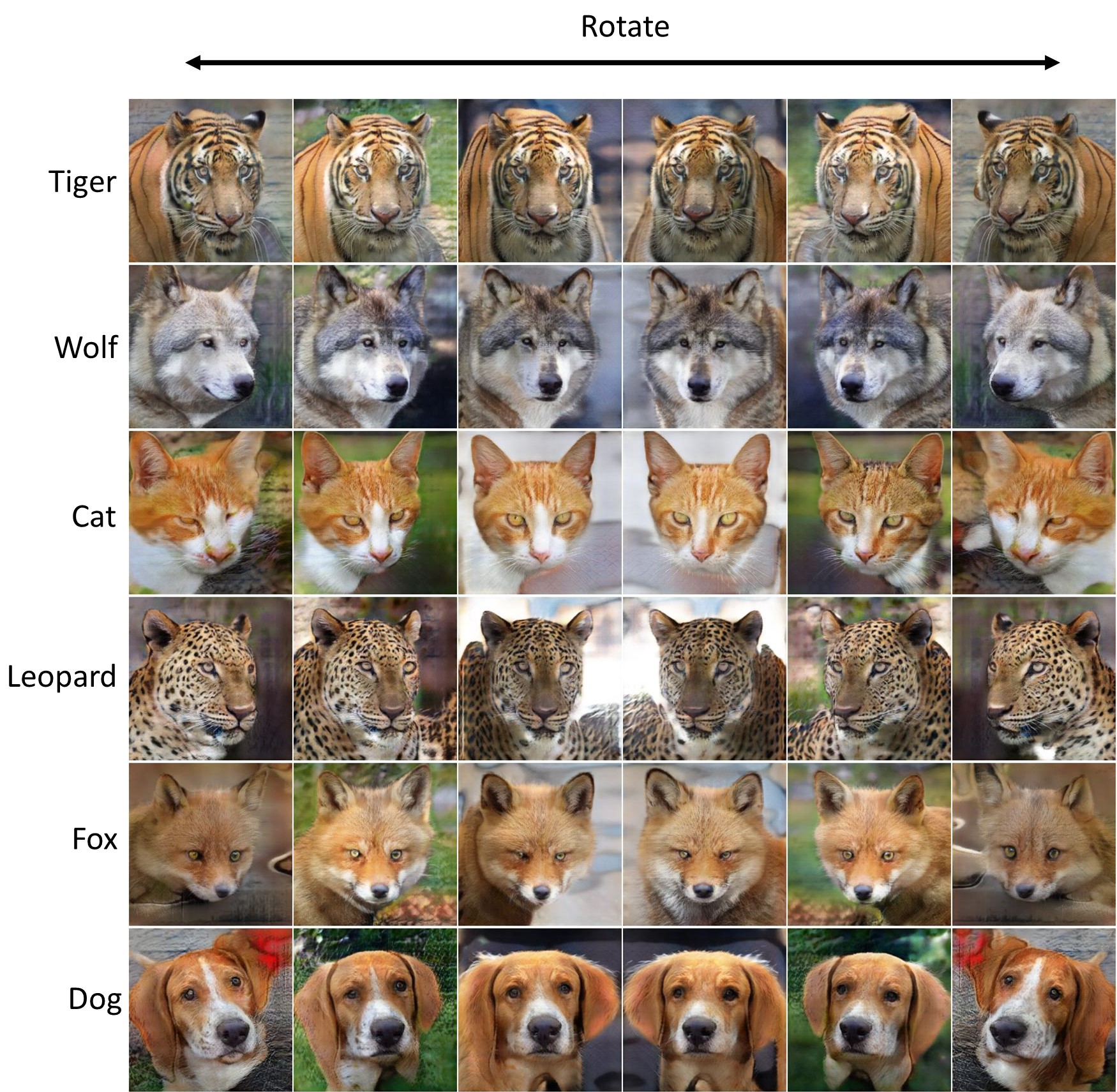}}
\caption{Synthesized images of each class of AFHQ by our model (with a $\textnormal{256}^{\textnormal{2}}$ resolution). A row displays a single object with different rotation input vectors. Note that the images of different classes are generated by a single model with different conditional input vectors. Our model can generate various views of different objects that conserves strong 3D-consistency.}
\label{figure_1}
\end{center}
\vskip -0.5in
\end{wrapfigure} 

Nevertheless, existing NeRF-based GANs cannot control image generation with conditional labels continuously, as they do not incorporate the necessary conditional information into the generator of the GAN. Conditioning NeRF-based GANs is crucial, as many industry applications demand precise manipulation during generation, such as selecting avatar details in the emerging metaverse. GIRAFFE \cite{niemeyer2021giraffe}, a NeRF-based generative model, can generate 3D-aware images without extra information, like camera location and direction, which conventional NeRF requires. Although GIRAFFE achieves impressive results, particularly in $360^{\circ}$ generation with real-world datasets compared to other state-of-the-art models like pi-GAN and Efficient Geometry-aware 3D Generative Adversarial Networks (EG3D) \cite{chan2021pi, EG3D}, it cannot disentangle various features \cite{bengio2013representation, locatello2019challenging, nguyen2020blockgan} needed to generate images with desired attributes.

Jo \etal{} \cite{jo2021cg} attempted to address this issue by providing condition information such as image types and texts. However, this approach is ineffective in representing condition intensity, as texts are ambiguous compared to numerical values, which offer a more intuitive means of setting intensity. Controlling condition intensity is essential, as most metaverse users want to customize their avatars with specific eye shapes or hair colors, which can be achieved by selecting numerical values that represent the intensity of each condition.

In this paper, we propose a novel method to tackle the task of 3D-aware conditional image generation. This task requires that conditional label values \cite{cgan, odena2017conditional, cgan_proj} control the features of generated images, as illustrated in Figure~\ref{figure_1}, and that these values be continuous to enable smooth changes in the corresponding condition intensities. To the best of our knowledge, this paper is the first to address this task.

We introduce our proposed method, Class-Continuous Conditional Generative Neural Radiance Field ($\textnormal{C}^{3}$G-NeRF), which focuses on conditional and continuous feature manipulation in 3D-aware image generation. Our backbone model is GIRAFFE, which enables $360^{\circ}$ generation without extra camera parameters, a core task in 3D-aware generation due to its wide range of applications. Although EG3D and pi-GAN outperform GIRAFFE in generation quality, they are not suitable for $360^{\circ}$ generation in real-world datasets. EG3D relies on extra camera parameters, making it unsuitable for learning real-world datasets without these parameters. Additionally, we validate the necessity of GIRAFFE by comparing its performance in generating a real-world car dataset \cite{ashrafi_2022} with pi-GAN using the Fréchet Inception Distance (FID) and Kernel Inception Distance (KID) metrics.



We observe that conditional GIRAFFE without residual modules struggles to learn data distributions with fine details, making residual modules essential for training conditional GIRAFFE. To address this issue, we incorporate residual modules \cite{he2016deep, he2016identity} into our model architecture to support training and improve image synthesis.

Addressing the challenge of generating multi-view instances in 3D-aware images, $\textnormal{C}^{3}$G-NeRF provides fine 3D-consistency across multiple views. We present results from three datasets, AFHQ \cite{choi2020stargan}, CelebA \cite{liu2015deep}, and Cars \cite{ashrafi_2022}, and demonstrate control over image synthesis through translation, rotation, and the addition of objects within a single image.

The contributions of this paper are as follows:
\begin{itemize}[noitemsep]
    \item We propose the $\textnormal{C}^{3}$G-NeRF model to address a novel task: conditional and continuous feature manipulation in 3D-aware image generation.
    \item We reduce training time and enhance performance by incorporating residual modules into the NeRF architecture.
    \item We showcase conditional and continuous feature manipulation in 3D-aware image generation using multiple datasets: AFHQ, CelebA, and Cars.
    \item Since our model can generate class-conditional 3D-aware images, we provide FID scores for each label in AFHQ and Cars datasets.
\end{itemize}

\section{Related Work}
\label{related_work}
\textbf{Implicit Neural Representation and Rendering:} The use of deep learning techniques \cite{lecun2015deep} to represent three-dimensional space has received considerable attention in recent years. Among the various approaches that have been proposed, implicit neural representations \cite{sitzmann2019scene, tulsiani2020implicit, rajeswar2020pix2shape} have shown particular promise. NeRFs have been proposed by combining an implicit neural representation with volume rendering \cite{drebin1988volume} to enable the synthesis of novel views that are not explicitly represented in the training data. As a result, NeRF is capable of synthesizing 3D-consistent images with fine details. 
However, one downside to using NeRFs is that they require highly constrained images for supervision during training. Another concern is that each instance of a NeRF can only represent a single object rather than multiple objects simultaneously. It has been proposed that generative NeRFs may be able to alleviate these problems.

\textbf{Generative NeRF:} There has been some recent progress in NeRF-based methods for generating 3D-aware images from 2D unconstrained image datasets. In these methods, generative models are trained to ensure continuous 3D geometric consistency. For instance, the GRAF \cite{schwarz2020graf} and pi-GAN \cite{chan2021pi} both proposed a generative NeRF, and showed promising results. GIRAFFE is another method that is more closely related to our work and improves on GRAF by separating an object from its background scene. However, none of these methods can control the conditional generation of images, which enables various feature manipulation in generated images. To address this issue, Jo \etal{} \cite{jo2021cg} proposed CG-NeRF. Specifically, CG-NeRF takes two types of conditional information. First, conditional data forms are used to translate one image into another; while our study aims to generate novel images directly from noise vectors instead. Second, CG-NeRF takes texts as conditions with CLIP (a natural language processing technique) \cite{radford2021learning}; whereas our model uses numerical conditions, which can be interpolated by varying values, allowing for continuous feature manipulation in image synthesis that represents intensity changes in relation to the given values.

\begin{figure*}[t]
\begin{center}
\centering {\includegraphics[width=\textwidth]{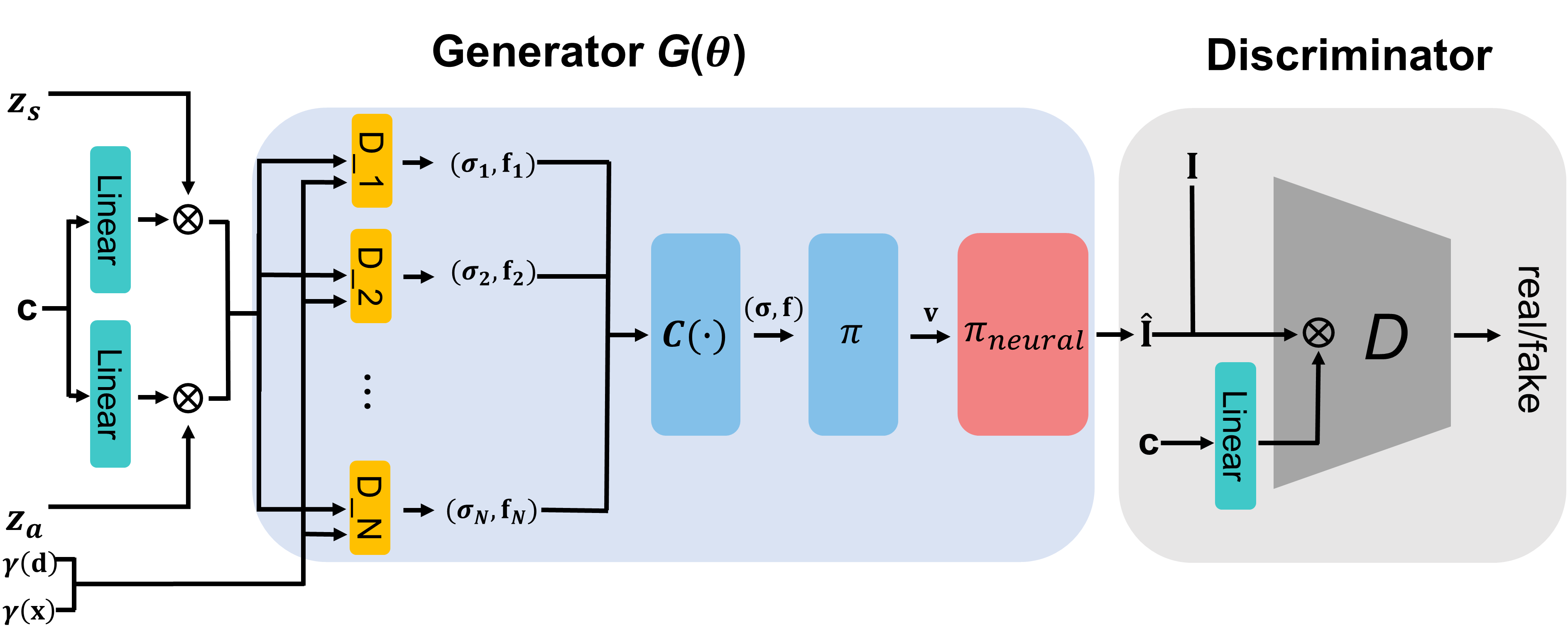}}
\caption{Overview of the proposed $\textnormal{C}^{3}$G-NeRF. Since our model is inspired by the architecture of GIRAFFE, our model generates $N-1$ objects and the background with $N$ decoders and a composition operator. D\_$i$ indicates $i$th decoder and $C(\cdot)$ represents the composition operator. The decoders take a 3D coordinate vectors of positional encoding $\gamma(\textbf{x})$ and viewing direction $\gamma(\textbf{d})$, where $\gamma$ indicates positional encoding functions. In addition, the decoders take conditional vectors $\textbf{c}$, which are encoded by linear layers, shape codes $\textbf{z}_{\textbf{s}}$, and appearance codes $\textbf{z}_{\textbf{a}}$. By compositing the outputs of each decoders with the composition operator $C(\cdot)$ and then volume-renders the result. Consequently, a composited feature vector $\textbf{v}$ is produced. The feature vector $\textbf{v}$ passes the neural rendering module $\pi_{neural}$. In this process, the generator $\textit{G}(\theta)$ synthesizes a fake image $\hat{\textbf{I}}$. The discriminator $\textit{D}$ takes a real image $\textbf{I}$ or the fake image $\hat{\textbf{I}}$ projected by the conditional labels $\textbf{c}$.}
\label{figure_2}
\end{center}
\vskip -0.4in
\end{figure*}

\section{Methods}
\label{methods}
Our goal is to make a framework for conditionally controllable 3D-aware image synthesis that guarantees representations of the intensity of conditions with continuous values. Given a labeled real-world 2D image dataset, the 3D-aware image generator \textit{G} takes conditions and camera pose, which are denoted by $\textbf{x}$ and $\textbf{d}$, respectively, and latent vectors for representing shape and appearance, i.e., $\textbf{z}_{\textbf{s}}$ and $\textbf{z}_{\textbf{a}}$. Then, \textbf{G} produces an image $\hat{\textbf{I}}$, corresponding to the input condition. At training time, real images from the dataset and $\hat{\textbf{I}}$ are directed to the discriminator \textbf{D}. 
Figure~\ref{figure_2} shows an overview of our model.

\subsection{Conditional Neural Radiance Fields} A conventional neural radiance field maps a 3D coordinate $\textbf{x} = (x, y, z)$ and viewing direction $\textbf{d} = (\theta, \phi)$ to a volume density $\sigma$ and a view-dependent RGB color value \text{R}, \text{G}, and \text{B} with fully-connected layers. However, several studies \cite{rahaman2019spectral} observed that deep learning techniques are difficult to represent high-frequency details, especially with low-dimensional inputs. To diversify the inputs, NeRFs introduced positional encoding \cite{tancik2020fourier} to the inputs, \textbf{x} and \textbf{d}, before the fully-connected layers:
\begin{equation}
    \gamma(p, L) = (\sin(2^{0}\pi p), \cos(2^{0}\pi p), \sin(2^{1}\pi p), \cos(2^{1}\pi p), \cdots, \sin(2^{\text{L}-1}\pi p), \cos(2^{\text{L}-1}\pi p)),
\end{equation}
where $\gamma(\cdot)$ indicates positional encoding, $L$ is the dimensionality of positional encoding, and $p$ is a scalar value as a component of \textbf{x} and \textbf{d}. To extend to generative neural radiance fields, shape and appearance codes, $\textbf{z}_{\textbf{s}}$ and $\textbf{z}_{\textbf{a}}$, are fed into the MLP as follows:
\begin{equation}
    (\gamma(\textbf{x}), \gamma(\textbf{d}), \textbf{z}_{\textbf{s}}, \textbf{z}_{\textbf{a}})\mapsto (\sigma, \text{R}, \text{G}, \text{B}).
\end{equation}

In our model, the generative neural radiance field was substituted to generative neural feature fields by extending the dimensionality of color \cite{niemeyer2021giraffe}, which was originally three-dimensional, to a feature space having a dimension of $M_{f}$ as:
\begin{equation}
    \text{h}_{\theta}: \mathbb{R}^{\text{L}_{\text{x}}} \times \mathbb{R}^{L_{\text{d}}} \times \mathbb{R}^{M_{\textbf{s}}} \times \mathbb{R}^{M_{\textbf{a}}} \mapsto \mathbb{R} \times \mathbb{R}^{M_{f}}, \hspace{0.5cm} 
    (\gamma(\textbf{x}), \gamma(\textbf{d}), \textbf{z}_{\textbf{s}}, \textbf{z}_{\textbf{a}})\mapsto (\sigma, \textbf{f}),
\end{equation}
where $\text{h}_{\theta}$ represents a generative neural feature field, $L_{\textbf{x}}$ and $L_{\textbf{d}}$ indicate dimensionalities of positional encoding output, and $M_{\textbf{s}}$ and $M_{\textbf{a}}$ are dimensionalities of latent encodings of the shape and appearance, respectively.

In this work, we project conditions to the latent vectors \cite{cgan_proj}, $\textbf{z}_{\textbf{s}}$ and $\textbf{z}_{\textbf{a}}$, by element-wise production. Before the projection, conditional vectors $\textbf{c}$ having a dimension of $M_{\textbf{c}}$ are encoded by a fully-connected layer to make the dimension the same as the dimensionalities of $\textbf{z}_\textbf{s}$ and $\textbf{z}_\textbf{a}$. The shape and appearance conditional encodings, $\textbf{c}_{\textbf{s}}$ and $\textbf{c}_{\textbf{a}}$, are constructed as
\begin{equation}
    \textbf{c}_{\textbf{s}} = L_{\textbf{s}}(\textbf{c}) * \textbf{z}_{\textbf{s}}, \quad \textbf{c}_{\textbf{a}} = L_{\textbf{a}}(\textbf{c}) * \textbf{z}_{\textbf{a}},
 \hspace{0.5cm}  L_{\textbf{s}}: \mathbb{R}^{M_{\textbf{c}}} \mapsto \mathbb{R}^{M_{\textbf{s}}}, \
    L_{\textbf{a}}: \mathbb{R}^{M_{\textbf{c}}} \mapsto \mathbb{R}^{M_{\textbf{a}}},
\end{equation}
where $L_{\textbf{s}}$ and $L_{\textbf{a}}$ are the encoding layers for the shape and appearance, respectively, and $*$ denotes element-wise multiplication. We employ these conditional projections as a replacement for conventional latent vectors in generative feature fields:
\begin{equation}
    h_{\theta}: \mathbb{R}^{L_\textbf{x}} \times \mathbb{R}^{L_\textbf{d}} \times \mathbb{R}^{M_\textbf{s}} \times \mathbb{R}^{M_\textbf{a}} \mapsto \mathbb{R} \times \mathbb{R}^{M_{f}},
   \hspace{0.5cm}  
     (\gamma(\textbf{x}), \gamma(\textbf{d})), \textbf{c}_{\textbf{s}}, \textbf{c}_\textbf{a} \mapsto (\sigma, \textbf{f}).
\end{equation}

\subsection{Scene Compositions}
As our model is motivated by GIRAFFE, it can separate scenes and individual objects with multiple feature vectors \cite{niemeyer2021giraffe}. 
Consequently, in the model, there are $N$ generative feature fields when $N - 1$ objects and a background scene exist in an image. To composite $N$ entities, the composition operator $C\left(\cdot\right)$ composites all feature fields from the $N$ entities. 
Each single entity $h_{\theta^{i}}^{i}$ outputs a volume density $\theta_{i}$ and a feature vector $\textbf{f}_{i}$. Following the method in GIRAFFE, we composite each component of entities with density-weighted mean-based composition. The mathematical expression of the composition operator can be represented as
\begin{align}
    C(\textbf{x}, \textbf{d}, \textbf{c}) = \Big(\sigma, \sum_{i=1}^{N}\frac{\sigma_{i}\textbf{f}_{i}}{\sigma}\Big),
\end{align}
where $\sigma = \sum_{i=1}^{N}\sigma_{i}$.

\subsection{Volume Rendering and Neural Rendering}
For scene rendering, 
our model volume-renders a camera ray $r(t) = \textbf{o} + t\textbf{d}$ to feature vectors \cite{niemeyer2021giraffe}, and subsequently, neural-renders the feature vectors to generate synthetic images. Our approach basically follows a discretized form of volume rendering methods used in NeRF \cite{mildenhall2021nerf}; nonetheless, detailed methods follow those of GIRAFFE to render features to images, which can be represented as
\begin{align}
    \textbf{v} = \sum_{j=1}^{N_{\textbf{s}}}T_{j}(1-e^{-\sigma_{j}\delta_{j}})\textit{f}_{j}, \hspace{0.5cm} \textnormal{where} \hspace{0.2cm}  T_{j} = \prod_{k=1}^{j-1}e^{-\sigma_{j}\theta_{j}},
\end{align}
where $\textbf{v}$ represents a final feature vector, $T$ denotes an accumulated transmittance along the cast ray, and $N_{\textbf{N}}$ is the number of sample points along a cast ray for an arbitrary camera pose $\xi$. By sampling points along the camera ray, we utilize a feature vector $\textit{f}_{j}$ and a density $\sigma_{j}$ corresponding to each point. Furthermore, the feature images have a $16^{2}$ resolution, i.e., $H_{V} \times W_{V}$ for cost-effectiveness.
Existing studies, including GRAF and NeRF \cite{schwarz2020graf, mildenhall2021nerf}, commonly adopted the volume rendering approach for 3D-to-2D projections. However, in our model, an additional 2D neural rendering network is required since the volume rendering composes feature images, not colored high-resolution images. The additional neural rendering network,
\begin{align}
    \pi_{neural}: \mathbb{R}^{H_{V}\times W_{V}\times M_{f}} \mapsto \mathbb{R}^{H\times W\times 3},
\end{align}
maps outputs of the volume rendering to a synthetic image, $\hat{\textbf{I}} \in \mathbb{R}^{H \times W\times 3}$, by upsampling the feature images. The architecture of neural rendering network is similar to a neural rendering operator in existing studies \cite{niemeyer2021giraffe}; however, we hire residual modules instead of conventional convolutional networks to enhance training speed and performance.

\subsection{Training Details}
At training time, we randomly sample latent vectors $\textbf{z}_\textbf{s}$ and $\textbf{s}_\textbf{a}$, as well as camera pose $\xi$ from prior distributions $p_{\textbf{s}}$, $p_{\textbf{a}}$, and $p_{\xi}$. Prior distributions $p_{\textbf{s}}$ and $p_{\textbf{a}}$ are defined as Gaussian distributions, and camera pose distribution $\xi$ are set to a uniform distribution. In the generator \textit{G}, we project conditional labels to latent vectors \cite{cgan_proj} $\textbf{z}_{\textbf{s}}$ and $\textbf{z}_{\textbf{a}}$. Similarly, conditional labels are projected to real images $\textbf{I}$ or synthesized images $\hat{\textbf{I}}$, in the discriminator $\textit{D}$. Real images $\textbf{I}$ are randomly sampled from the training dataset, which follows distribution $p_{\textbf{I}}$. We use a GAN loss with R1 gradient penalty \cite{mescheder2018training} as follows:
\begin{align}
    \mathcal{L}(\textit{G}, \textit{D}) =& \mathbb{E}_{\textbf{z}_{\textbf{s}}\sim p_{\textbf{s}}, \textbf{z}_{\textbf{a}}\sim p_{\textbf{a}}, \xi\sim p_{\xi}}\big[
    -\log(\textit{D}(\textit{G}(\textbf{z}_{\textbf{s}},\textbf{z}_{\textbf{a}},\xi,\textbf{c})))\big] \\ \nonumber 
    &+ \mathbb{E}_{\textbf{I}\sim p_{\textbf{I}}}
    \big[-\log(1-\textit{D}(\textbf{I})) + \lambda \|\nabla \textit{D}(\textbf{I})\|^{2}\big].
\end{align}
We train the generator and discriminator of the proposed model by competing for a zero-sum game with the above loss function. We use the RMSprop optimizer \cite{rmsprop} with a learning rate of $3\times 10^{-4}$ and $1\times 10^{-4}$ for the generator and the discriminator, respectively. The model is trained on one A6000 GPU with 48GB memory with a batch size of 32. We set $M_{f} = 128$ for a $64^{2}$ resolution and a $128^{2}$ resolution, while we set $M_{f} = 256$ for a $256^{2}$ resolution. We utilize ReLU activations \cite{relu} as activation functions used in our model, except for the final layer in the neural renderer, which uses a sigmoid function \cite{ramachandran2017searching}.


\subsection{Accelerating Training with Residual Modules}
We observe elaborate conditional generation with GIRAFFE is not feasible without residual modules \cite{he2016deep, mescheder2018training}. In the experiments of this study, we demonstrate that conditional generation with plain networks shows lagging performance. We argue that this result is because conditionally projected latent vectors are too far from the discriminator, which causes gradient vanishing. Therefore, residual modules support conveying the information to conditionally projected latent vectors from the discriminator. Moreover, the range of varying latent vectors expands since we projected conditional labels to latent vectors, which is challenging to learn with plain networks. We apply residual modules to our model in the decoder, neural rendering network, and discriminator. We validate the efficiency of this contribution in the Section \ref{experiments}.

\section{Experiments}
\label{experiments}
In this section, we evaluate our $\textnormal{C}^{3}$G-NeRF on three real-world datasets: CelebA \cite{liu2015deep}, AFHQ \cite{choi2020stargan}, and Cars \cite{ashrafi_2022}. 
We first evaluate the conditionally controlling 3D-consistent image generations. We then evaluate the quality of generation by FIDs \cite{ttur}. 
Finally, we include an evaluation of residual modules to validate the efficiency of residual modules adopted in our model.

\subsection{Controllable Features in 3D Object Generation}

\begin{figure}
\begin{tabular}{cc}
\bmvaHangBox{\includegraphics[width=8cm]{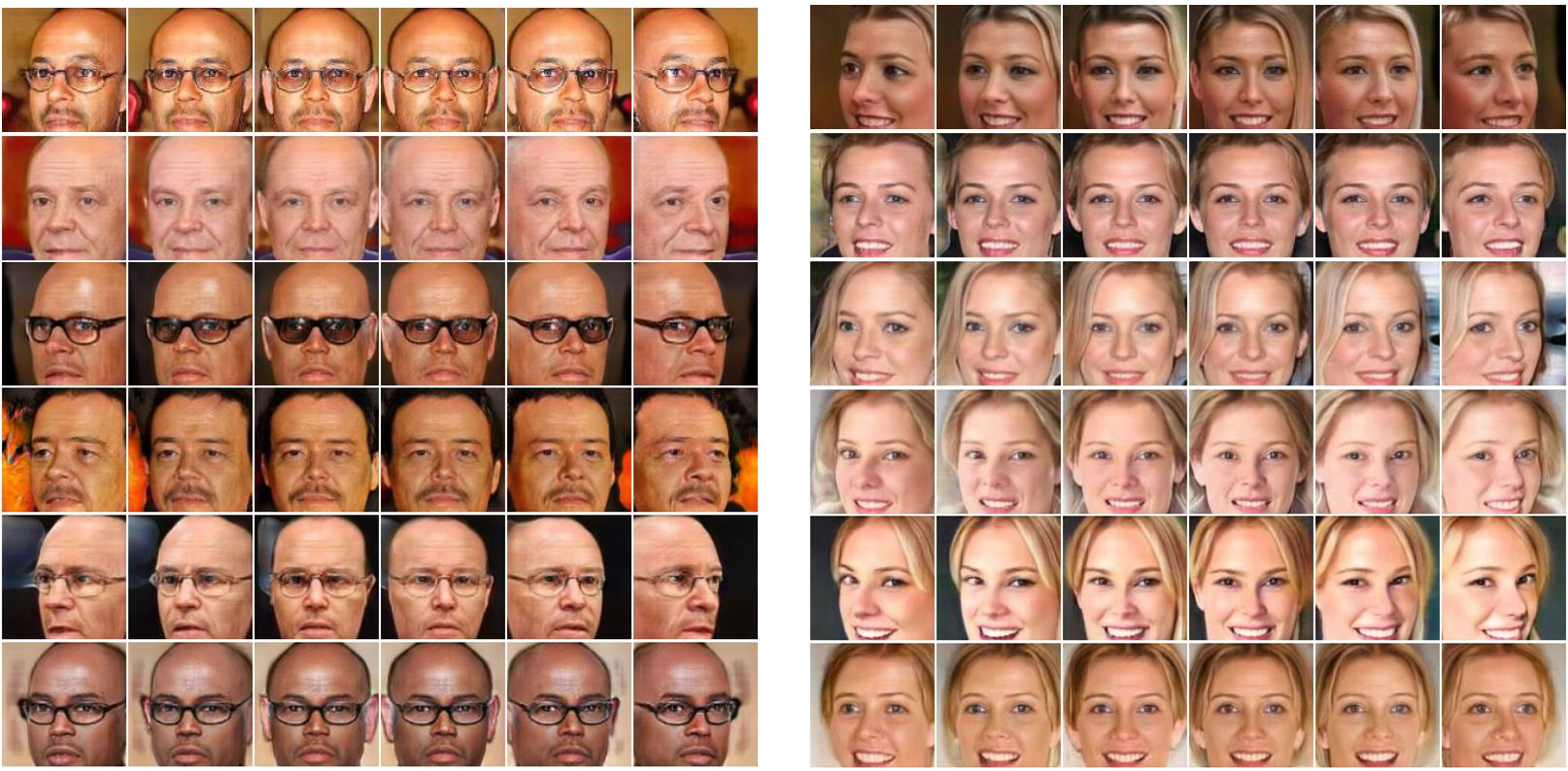}}&
\bmvaHangBox{\includegraphics[width=4cm]{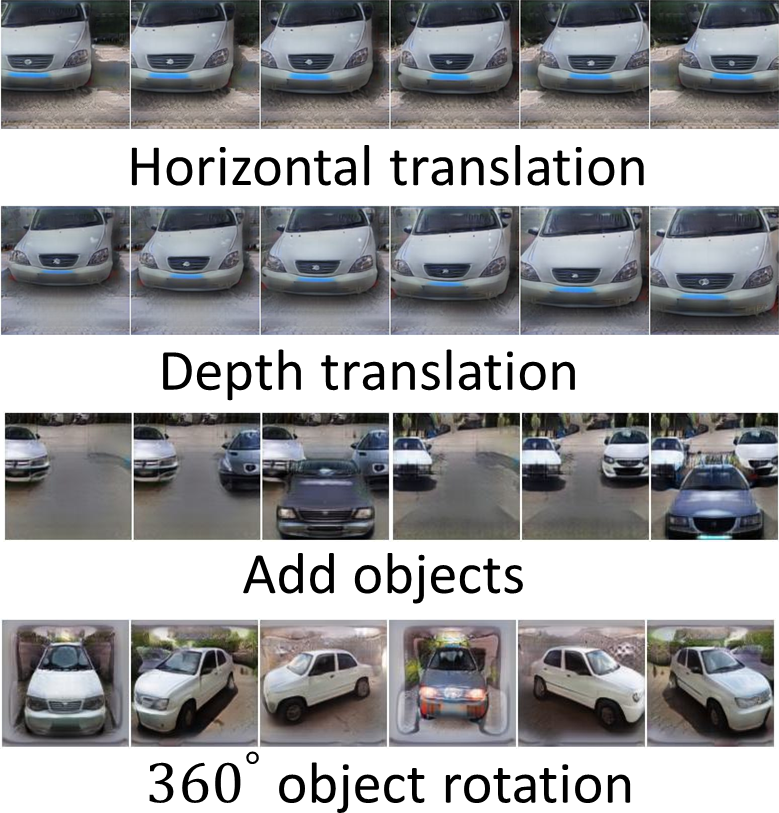}}\\
(a)&(b)
\end{tabular}
\caption{Class conditional synthetic object rotation generated by $\textnormal{C}^{3}$G-NeRF trained with CelebA and Cars. In (a), each row represents a single object of CelebA with the same latent vectors. Each column indicates rotation angles. In the left figure of (a), we fixed the input conditions as a bald man, whereas we fixed the conditions as a blonde smiling woman in the right figure of (a). In (b), by controlling the horizontal and depth translation, the disentanglement of the objects and background are shown in Horizontal translation and Depth translation. After training with unstructured 2D images with a single object, we can generate $N-1$ objects in one scene by replicating $N$ decoders as in Add objects. 
  All images have a resolution of $128^2$. Using $\textnormal{C}^{3}$G-NeRF, 3D-consistent image generation is successful under the given conditions.
}
\label{figure_3}
\vskip -0.2in
\end{figure}

The model evaluation involves performing object rotations, horizontal translations, depth translations, and object additions. The results of this evaluation are presented in Figure~\ref{figure_1} and \ref{figure_3}. These figures demonstrate that $\textnormal{C}^{3}$G-NeRF effectively learns 3D-consistency for AFHQ, CelebA, and Cars datasets, respectively, by preserving spatial consistency despite the introduced transformations. Additionally, the model is observed to successfully capture the conditional input features and generate images accordingly.

Furthermore, in Figure~\ref{figure_3}, we assess $\textnormal{C}^{3}$G-NeRF using out-of-distribution images with translating depth and horizontal at test time. This means $\textnormal{C}^{3}$G-NeRF can generate beyond the distribution of training images by using extended rotation and transition values over the training sets. Moreover, we can finely control each object and scene in the generated images by $\textnormal{C}^{3}$G-NeRF; for instance, while adding the objects in a scene, each object can be controlled by translating and rotating in 3D space.

\subsection{Continuously Controllable Features in 3D Object Generation}

\begin{figure}[t]
\begin{tabular}{cc}
\bmvaHangBox{\includegraphics[width=4.4cm]{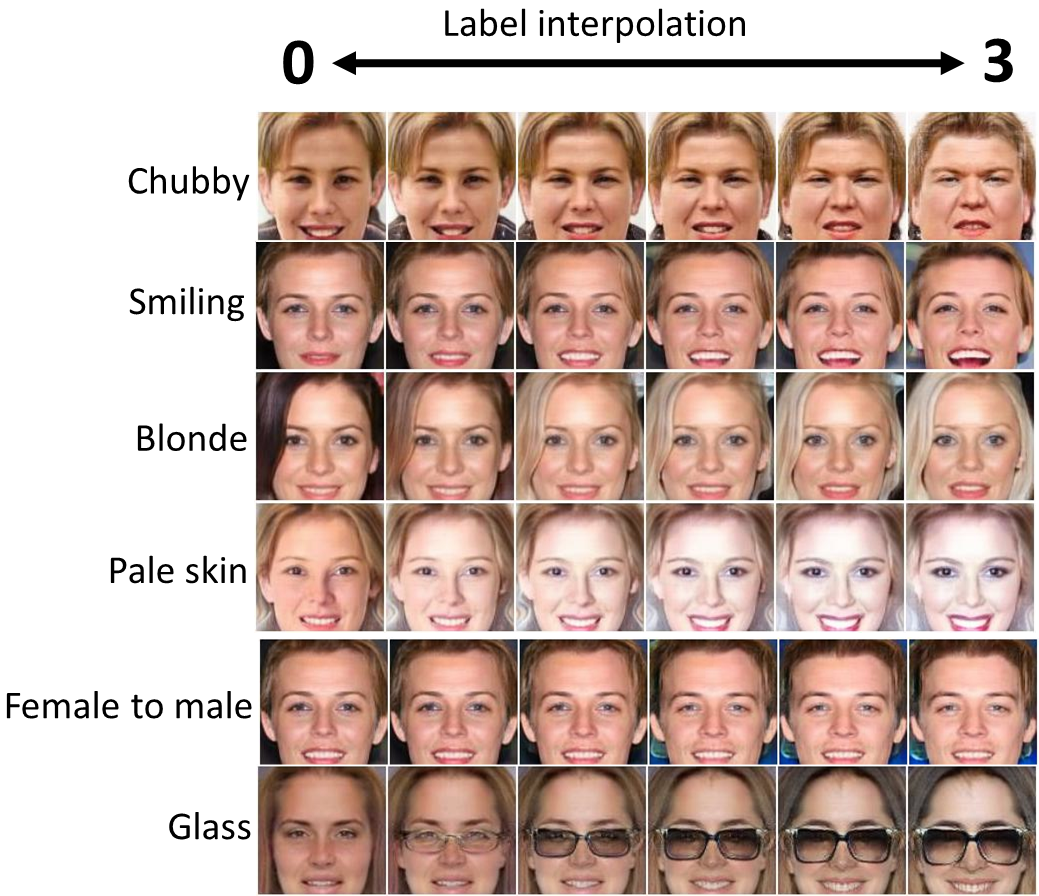}}&
\bmvaHangBox{\includegraphics[width=7.7cm]{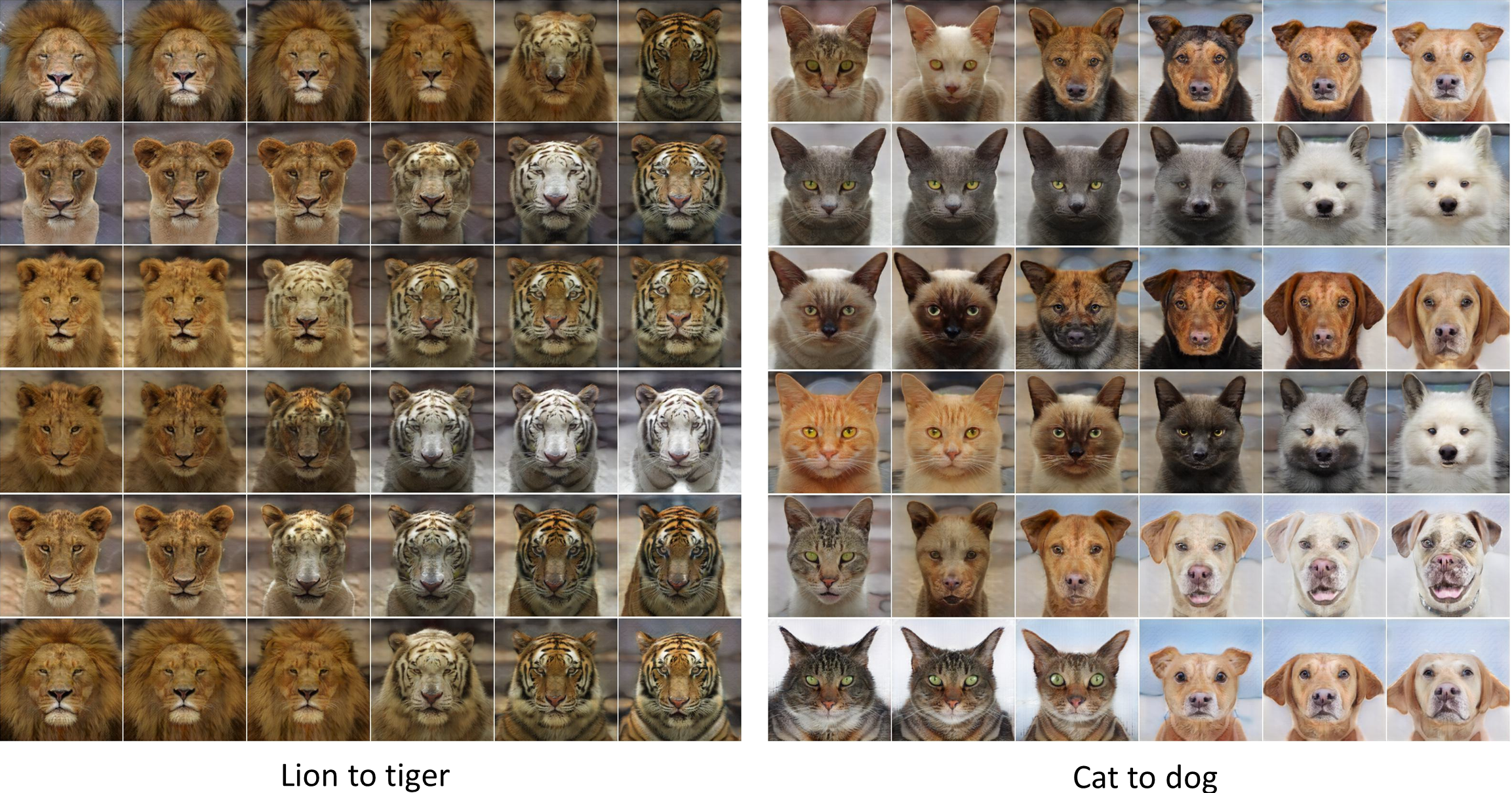}}\\
(a)&(b)
\end{tabular}
\caption{Interpolation and extrapolation on conditional input values with CelebA and AFHQ. Each row and column represent the same latent vectors (identical object) of AFHQ and the same class-conditional values, respectively. In (a), we present the conditional results according to conditional input values with the range of zero to three. Note that, in training time, the features are trained only with the two values of zero and one, which indicate the existence of the corresponding feature. The features in the face images with interpolated and extrapolated input values smoothly change, which means the continuous conditional learning is adequately progressed. By interpolating the values of each class, features of each category coexist at the intermediate state of class-conditional values.
}
\label{figure_4}
\vskip -0.2in
\end{figure}

We demonstrate our model's ability to manipulate individual conditions by interpolating 40 conditional binary values in the CelebA dataset, such as chubby, smiling, blonde, and pale skin. Although the training procedure has a range of zero to one due to binary encoding in the CelebA dataset, we expand the test procedure range to zero to three, as illustrated in Figure~\ref{figure_4}. This experiment assesses whether each facial feature maps to the generator's input label. With various conditional label values, $\textnormal{C}^{3}$G-NeRF exhibits superior performance in interpolation and extrapolation for each condition. For instance, chubby and smiling conditions \cite{liu2015deep} change smoothly according to conditional values, regardless of the extrapolation range. This result highlights our model's capacity to generate out-of-distribution images, such as exaggerated features with high input label values exceeding one.


We also examine non-characteristic (implicit) labels corresponding to AFHQ classes, which are more challenging to learn due to less obvious image features. We generate inter-class images with interpolated class-conditional input values, as depicted in Figure~\ref{figure_4}. We observe that features of each class coexist at intermediate class-conditional values, indicating that feature manipulation can be applied to implicit class labels.



\begin{table}[t!]
\vskip 0.1in
\begin{small}
\begin{center}
\begin{sc}
\resizebox{\columnwidth}{!}
{
\begin{tabular}{c|ccc|cc|ccc}
\hline\hline
\multirow{2}{*}{Model} & \multicolumn{3}{c|}{AFHQ (FID)} & \multicolumn{2}{c|}{CelebA(FID)} & \multicolumn{3}{c}{Cars(FID)}\\
\cline{2-9}
      & $64^{2}$ & $128^{2}$ & $256^{2}$ & $64^{2}$ & $128^{2}$ & $64^{2}$ &$128^{2}$ & $256^{2}$ \\
\hline
 Baseline & 212.74 & 226.51 & 239.01 & 55.90 & 83.89 & 228.66 & 244.55 & 266.51 \\
 Ours & \textbf{26.72} & \textbf{25.79} & \textbf{28.58} & \textbf{5.60} & \textbf{7.64} & \textbf{44.79} & \textbf{43.29} & \textbf{31.63} \\
 
\hline
\hline
\multirow{2}{*}{Model} & \multicolumn{3}{c|}{AFHQ (KID)} & \multicolumn{2}{c|}{CelebA(KID)} & \multicolumn{3}{c}{Cars(KID)}\\
\cline{2-9}
      & $64^{2}$ & $128^{2}$ & $256^{2}$ & $64^{2}$ & $128^{2}$ & $64^{2}$ &$128^{2}$ & $256^{2}$ \\
\hline
 Baseline & 0.200 & 0.352 & 0.323 & 0.119 & 0.296 & 0.096 & 0.166 & 0.303 \\
 Ours & \textbf{0.077} & \textbf{0.059} & \textbf{0.051} & \textbf{0.046} & \textbf{0.023} & \textbf{0.057} & \textbf{0.039} & \textbf{0.043} \\
\hline
\hline
\end{tabular}}
\end{sc}

\end{center}
\vskip -0.25in
\caption{Quantitative comparison with FIDs ($\downarrow$) and KIDs ($\downarrow$) with three datasets. The baseline is set to the conditional GIRAFFE with plain networks. The $64^{2}$, $128^{2}$, and $256^{2}$ are the image resolutions of the generated images and real images.}
\label{table_1}
\end{small}
\vskip -0.1in
\end{table}

\begin{table}[t!]
\begin{small}
\begin{center}

\begin{sc}
\begin{tabular}{  >{\centering\arraybackslash}m{2.4cm} | >{\centering\arraybackslash}m{1cm} >{\centering\arraybackslash}m{1cm} > {\centering\arraybackslash}m{1cm} | >{\centering\arraybackslash}m{1.4cm} | >{\centering\arraybackslash}m{1cm} >{\centering\arraybackslash}m{1cm} >{\centering\arraybackslash}m{1cm}} 
\hline\hline
\multirow{2}{*}{Category} & \multicolumn{3}{c|}{CARS} & \multirow{2}{*}{Category} & \multicolumn{3}{c}{AFHQ}\\
\cline{2-4} \cline{6-8}\
      & $64^{2}$ & $128^{2}$ & $256^{2}$ & & $64^{2}$ & $128^{2}$ & $256^{2}$\\

 \hline
 Peykan & 73.54 & 67.47 & 68.57 & cat & 10.38 & 13.65 & 15.48 \\
 Quik & 58.73 & 62.71 & 49.64 & cat & 10.38 & 13.65 & 15.48 \\
 Samand & 54.63 & 50.53 & 61.01 & dog & 31.64 & 43.37 & 51.73 \\
 Peugeot-Pars & 65.34 & 66.27 & 46.45 & leopard & 17.20 & 14.98 & 13.42\\
 Peugeot-207i& 71.17 &71.06 & 52.99 & fox & 25.97 & 28.01 & 22.50\\
 Pride-111 & 67.91 & 62.28 & 68.84 & lion & 8.76 & 12.20 & 7.61\\
 Pride-131 & 68.70 & 57.70 & 65.43 & tiger & 12.78 & 8.96 & 5.93\\
 Tiba2 & 60.80 & 61.79 & 55.57 & wolf & 28.39 & 30.82 & 14.85\\
 
\cline{5-8} Renault-L90 & 67.79 & 65.51 & 76.84\\
 Nissan-Zamiad & 78.17 & 88.83 & 133.77\\
 Peugeot-206 & 63.53 & 61.55 & 128.10\\
 Peugeot-405 & 71.79 & 66.15 & 71.45\\
 Mazda-2000 & 71.75 & 77.23 & 104.53\\

\hline
\hline
\end{tabular}
\end{sc}

\end{center}
\vskip -0.25in
\caption{Quantitative comparison with FIDs ($\downarrow$) with each class of AFHQ and Cars. The $64^{2}$, $128^{2}$, and $256^{2}$ are the image resolutions of the generated images and real images.}\label{table_2}
\end{small}

\vskip -0.25in
\end{table}


\subsection{Quantitative Evaluation}

We evaluate image quality using a conventional method with KIDs and FIDs, comparing 20,000 randomly sampled real and generated images. Assessments demonstrate generation quality for each dataset and label, evaluating images generated with random conditional inputs across different resolutions and datasets.

Table~\ref{table_1} shows that $\textnormal{C}^{3}$G-NeRF achieves impressive FIDs and KIDs in conditional 3D-consistent generation, outperforming conditional GIRAFFE regardless of resolution and indicating robustness in high resolutions. The conditional GIRAFFE exhibits high FIDs, suggesting training issues. For instance, in CelebA face image generation at $128^{2}$ resolution, $\textnormal{C}^{3}$G-NeRF attains an FID of 7.64, reducing the value by $90.9\%$ compared to the baseline.

\begin{wraptable}{R}{7.5cm}
\begin{small}
\begin{center}
\vskip -0.25in
\begin{tabular}{>{\centering\arraybackslash}m{1.2cm} | >{\centering\arraybackslash}m{1.0cm}>{\centering\arraybackslash}m{1.0cm}|>{\centering\arraybackslash}m{1.0cm}>{\centering\arraybackslash}m{1.0cm}}
\hline\hline
\multirow{2}{*}{Model} & \multicolumn{2}{c|}{Cars (KID)} & \multicolumn{2}{c}{Cars (FID)}\\
\cline{2-5}
      & $64^{2}$ & $128^{2}$ & $64^{2}$ & $128^{2}$\\
\hline
 pi-GAN & 0.105 & 0.083 & 137.34 & 104.51\\
 Ours & \textbf{0.057} & \textbf{0.039} & \textbf{44.79} & \textbf{43.29}\\
\hline
\hline
\end{tabular}
\end{center}
\vskip -0.1in
\caption{Quantitative comparison with FIDs ($\downarrow$) and KIDs ($\downarrow$) on Cars dataset for 360$^{\circ}$ generation at resolutions $64^{2}$ and $128^{2}$, with comparison to pi-GAN.}
\label{table_3}
\vskip -0.1in
\end{small}
\end{wraptable}

Table~\ref{table_2} presents FIDs for each label, reflecting the conditional disentanglement achieved by $\textnormal{C}^{3}$G-NeRF. FIDs for individual labels are similar to or better than those for all labels combined, implying that $\textnormal{C}^{3}$G-NeRF effectively learns features for each condition across datasets. Moreover, even with few examples (e.g., foxes, lions, tigers in AFHQ), comparable FIDs indicate that $\textnormal{C}^{3}$G-NeRF excels at learning from data, irrespective of its unposed distribution.

Table~\ref{table_3} shows that our model outperforms pi-GAN on Cars dataset, effectively disentangling class conditions. Generally, conditional models underperform unconditional ones, rendering pi-GAN unsuitable for $360^{\circ}$ generation. Additionally, EG3D struggles with Cars dataset as it needs camera positions corresponding to samples while EG3D can obtain camera positions using an independent algorithm for facial datasts \cite{deng2019accurate, cat_hip}. Thus, we adopt GIRAFFE for the generality in $360^{\circ}$ generation with real-world datasets.



\subsection{Evaluation of Residual Modules}
We assess image quality generated by $\textnormal{C}^{3}$G-NeRFs with residual modules, comparing it to conditional GIRAFFE with a plain network. This experiment emphasizes the impact of incorporating residual modules \cite{he2016deep, mescheder2018training} in the $\textnormal{C}^{3}$G-NeRF architecture. FID and KID comparisons across three datasets (Table~\ref{table_1}) show that $\textnormal{C}^{3}$G-NeRF significantly outperforms conditional GIRAFFE with plain networks, supporting our hypothesis that residual modules enhance generation quality and are crucial for conditional 3D-aware generation. Residual modules aid gradient flow from the discriminator's output to the conditional inputs of the generator, facilitating conditional information learning. Qualitative comparisons are shown with the Figure 12, which is included in the Supplementary Materials.

\section{Conclusions}
We presented a novel model, $\textnormal{C}^{3}$G-NeRF, for conditional and continuous feature manipulation in 3D-aware image generation. Our approach projects conditional labels with encoding layers onto the generator's latent vectors and an intermediate discriminator layer to disentangle dataset features. $\textnormal{C}^{3}$G-NeRF leverages residual modules to optimize 3D-aware conditional training. Interpolating and extrapolating conditional input values, we achieved precise 3D-consistent image generation and feature manipulation. 

\section{Acknowledgement}
This research was supported by the National Research Foundation of Korea (NRF) grant funded by the Korea government (MSIT) (No. RS-2023-00251528).

\bibliography{egbib}
\end{document}